\documentclass{article}
\usepackage{acl2015}
\usepackage{times}
\usepackage{url}
\usepackage{latexsym}
\usepackage{graphicx}
\graphicspath{ {files/} }
\title{Classifying medical notes into standard disease codes}

\author{
  Amitabha Karmakar \\
  University of California \\
  Berkeley, CA \\ 
  { amitkarmakar@berkeley.edu } \\} 
  
\begin{document}
\maketitle
\begin{abstract}
We investigate the automatic classification of patient discharge notes into standard disease labels. We find that Convolutional Neural Networks with Attention outperform previous algorithms used in this task, and suggest further areas for improvement.
\end{abstract}

\section{Introduction}
Electronic Health Records (EHRs) have grown significantly over the years and now include an unprecedented amount and variety of patient information, including demographics, vital sign measurements, laboratory test results, prescriptions, procedures performed, digitized notes, imaging reports, mortality etc. They usually contain both structured data (e.g. admission dates) as well as unstructured data (e.g. notes written by doctors).

Provided it can be processed, the information in these records - especially the unstructured data - holds the promise of new medical insights and improved medical care, such as faster detection of epidemics, identification of symptoms, personalized treatment, or a more detailed understanding of treatment outcomes.

One such gains is a more automated and accurate way to report diseases. Since 1967, the World Health Organization (WHO) has developed an International Classification of Diseases (ICD) to ``monitor the incidence and prevalence of diseases, observe reimbursements and resource allocation trends, and keep track of safety and quality guidelines''\footnote{http://www.who.int/classifications/icd/en/}. Currently this ICD labeling is done manually by administrative personnel based on definitions and is subject to interpretation and errors\footnote{See recent articles on opiod overdose statistics in the US https://tinyurl.com/y8zc3xcq}.

In this paper, we focus our efforts on the automatic labeling of discharge notes from the MIMIC\footnote{Medical Information Mart for Intensive Care https://mimic.physionet.org} Database into ICD codes. This public database of EHRs contains data points on about 41,000 patients from an intensive care units between 2001 and 2012, including notes on close ot 53,000 admissions. MIMIC has already been proven valuable for efforts similar to ours, which will make comparisons more accurate.

\section{Background}

The problem of assigning ICD codes automatically to discharge summaries has previously been studied. Among the most recent research publications, we can distinguish two types of efforts: papers trying to predict the ICD codes in all their complexity, and those more numerous who focus on a smaller domain.   

{\bf Full ICD codes}: Perotte et al.~\shortcite{Perotte:14} used the MIMIC II dataset to predict the original ICD codes. They experimented with two approaches: one that treats each ICD9 code independently of each other (flat classifier), and one that leverages the hierarchical nature of ICD9 codes into its modeling (hierarchy-based classifier). They used a novel evaluation metrics, which reflected the distances among source ICD9 tree and predicted codes and their locations in the ICD9 tree. They found that the hierarchy-based classifier outperformed the flat classifier.

{\bf Simplified ICD codes}: Other researchers focused their efforts on a smaller number of ICD codes, and found their best results using Convolutional Neural Networks (CNNs). Gerhman et al.~\shortcite{Gehrmann:17} relabeled 1.6K clinical notes from MIMIC III using their own 10 labels. They find that CNNs outperform other approaches based on n-gram models, and Natural Entity Recognition (NER).\footnote{Using cTAKES http://ctakes.apache.org/} 

{\bf Our Approach}: In this paper, we focused on improving techniques applied to the simplified ICD code problem. While CNNs seem well suited, some characteristics of discharge notes raise possibilities for other approaches. 

Medical notes describe a temporal sequence of events and tests, to which CNNs are oblivious, on the contrary to Long Short-Term Memory (LSTM) models. Additionally, notes are long, with an average length of around 1500 words. Because of this large context, we also explored Attention models which do not seem to have been applied to this problem domain before.

Last, we briefly investigated how the two approaches (full and simplified ICD codes) could be reunited by adapting the training metric.

\section{Methods}

We focused on the classification of hospital admission discharge notes into ICD-9 codes, using the MIMIC III database \cite{Johnson:16} for comparison purposes. 

We can broadly break our approach to this multi-label multi-class problem into the steps detailed below: output labeling, input preprocessing, training and output metrics, and algorithms.

\subsection{Output labeling}
\label{ss:labeling}
The ICD-9 nomenclature applied by MIMIC III contains about 14,000 numerical codes representing all possible diagnoses and procedures\footnote{ICD-10 (current version) has around 68,000 labels}. Out of those 14,000, 5,932 distinct codes are used to describe the 52,696 hospital admissions of the database, with 1,112 codes appearing only once.

\begin{figure}[h!]
  \caption{Sample ICD9 path\footnote{https://cs224d.stanford.edu/reports/lukelefebure.pdf}}
  \label{fig:icd9_hierarchy}
  \centering
  \includegraphics[width=0.5\textwidth]{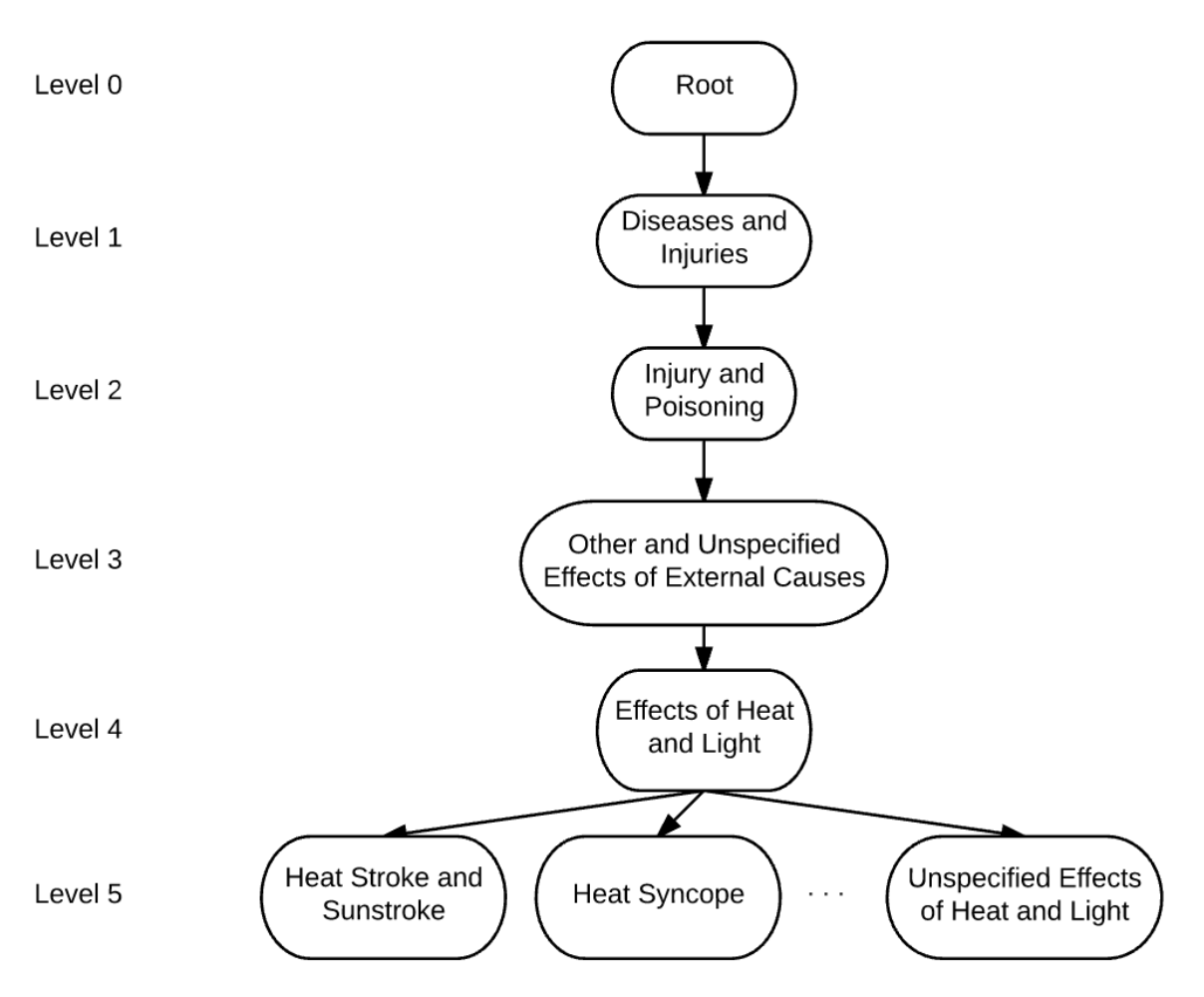}
\end{figure}

This creates an issue for classification algorithms since many codes would need to be predicted with few or no example in the training set. Fortunately, the ICD codes are organized in a hierarchical tree, see Figure~\ref{fig:icd9_hierarchy}.

\begin{figure*}[!h]
  \caption{Penetration of top 20 Level 5 codes (left) and all Level 1 codes (right)}
  \label{fig:merged}
  \centering
  \includegraphics[width=1\textwidth]{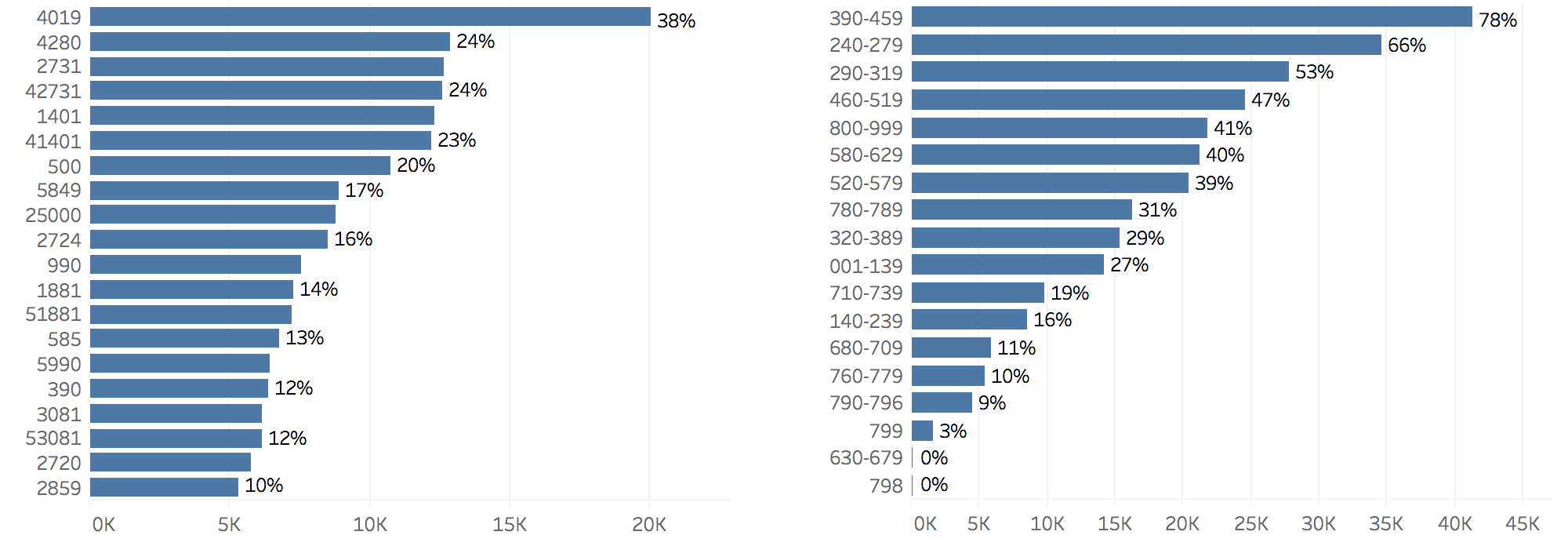}
\end{figure*}

As a result, we identified 3 mains methods to deal with the high number of classes:

\begin{itemize}
\item Restrict the labels to the most common Level 5 codes, method used by some project reports.\footnote{https://cs224d.stanford.edu/reports/lukelefebure.pdf} We start by selecting the 20 most common codes (see Figure~\ref{fig:merged})
\item Relabel all codes into a smaller class of codes. This approach was done manually by \cite{Gehrmann:17}. Here, we take advantage of the ICD hierarchy, and simply relabel notes into the 17 nodes of depth 1.\footnote{Excluding 798 which appears only once}
\item A third possible approach - to explore in further work - would be to keep all labels, but use a "hierarchical metric", i.e. discounting errors if labels are in the same ICD branch.
\end{itemize}

Intuitively, we can expect the second approach to perform better, since i) the codes  represent very different realities, whereas common codes may be related, and ii) the distribution is less balanced. An additional benefit is that we have access to the full dataset for training (53K) instead of just a subset if we had re annotated the dataset manually or if we take the most common codes (46k).

The third untested approach would allow to keep all codes intact, and hence be more precise in the labeling.

\subsection{Note preprocessing}
\label{ss:preprocessing}

The database presents multiple clinical notes categories including things like “Radiology”, “Nutrition”, “Pharmacy”, or “Social Work”. Here, we focus on “Discharge Summaries”,\footnote{There are 2 types of discharge summaries, reports and addendum, we focused on the reports} which already provide a synthesis of different aspects.

To process those notes, we go through relatively common steps that we summarize briefly here: we put words in lower case, remove most special characters, separate contractions, canonize numbers, and tokenized the resulting words.

The results is a vectorized set of notes, which can reach 10,924 words. Since some of our algorithms require a fixed length input, we truncate and pad the notes so that the output has a length of 5,000 words. This is done without loss of generality, since 99.5\% of the notes meet this criteria.

\begin{figure}[h!]
  \caption{Original distribution of note length}
  \label{fig:length_notes}
  \centering
  \includegraphics[width=0.5\textwidth]{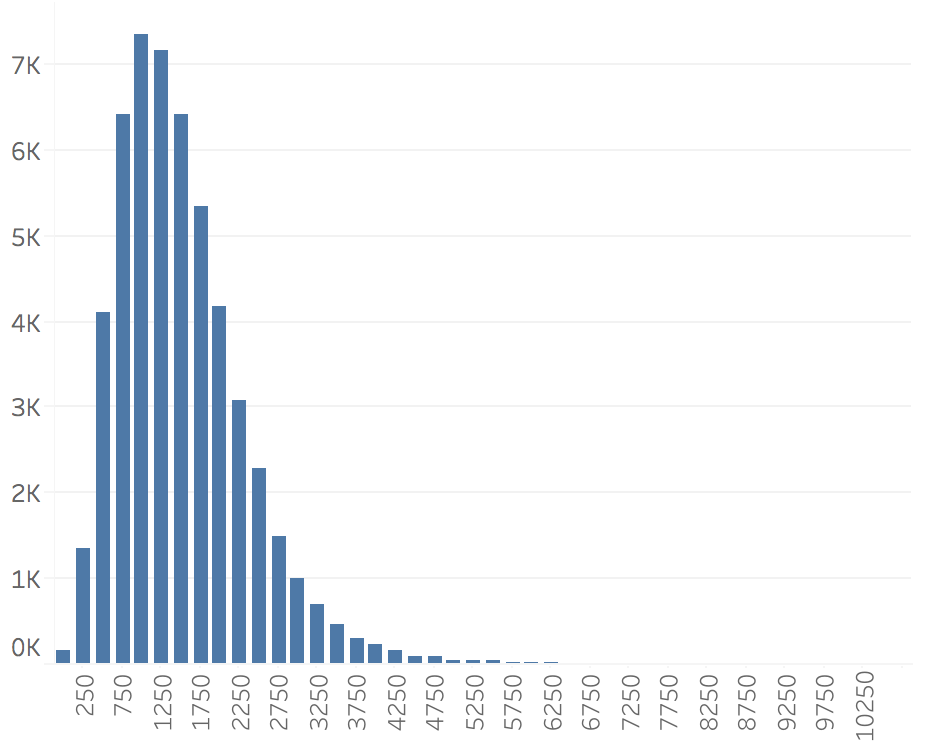}
\end{figure} 

\subsection{Embedding}
\label{ss:embedding}

Unfortunately, even after the previous steps, the wording is still not standardized. Like some unpublished papers, we can see at least 13 ways that write hypercholesterolemia, for instance.\footnote{https://cs224d.stanford.edu/reports/priyanka.pdf}

One way to solve this issue would be to use Named Entity Recognition (NER). Some implementations exist which are tailored to the medical realm, such as Apache cTAKES or MetaMap. However, previous papers \cite{Gehrmann:17} find that embeddings perform better, trusting embeddings' ability to make ``misspellings, synonyms and
abbreviations of an original word learn similar embeddings''. Therefore we used trainable embeddings, sometimes pre-trained with the Glove algorithm on Wiki\footnote{https://nlp.stanford.edu/projects/glove/} or on the MIMIC notes to account for the vocabulary specificity.

Note that some papers such as \cite{Perotte:14} use TF-IDF, either to restrict the original vocabulary size or to transform notes into continuous components. Here since we use embeddings of size 100, we can keep our original vocabulary of 60,619 with limited impact to our calculation time. 

\subsection{Training Loss Function}

For multi-label classifications like this one, an approach is to convert the problem into single binary classification tasks. This would not work for ICD-9 codes since the ones assigned to a clinical-note may not be independent (some medical conditions are correlated).

An early procedure for multi-label classification using NNs was BP-MLL which uses a novel pairwise ranking loss function for training \cite{Zhang:06}, but later research found that cross entropy produces better results \cite{Nam:13}. In this work, we use the latter.

\subsection{Algorithms}

For all of our models, we used standard software and evaluation methods.\footnote{Data is split between training (70 \%), validation (15\%) and test (15\%) sets. Models were implemented using Tensorflow and Keras. Training optimizer used was Adam. Our models are using L2 regularizations and Dropouts, which proved its efficiency.Default parameters were used.}

\subsubsection{Baseline and Linear Models}
Our Baseline model simply predicts the 4 most common ICD-9 codes for each clinical note.
This performs better than more complicated alternatives, for example Gehrmann \bgroup et al.\egroup~\shortcite{Gehrmann:17} used a 3-gram Logistic Regression with relatively poor results (Table~\ref{table:CNN_5k}).

\subsubsection{CNN}

CNNs have been used for image multi-label classification. Although the invariances are different between an image and a text, this sounds similar to our problem. 

This work implements a CNN for text classification replicating the architecture presented by Kim ~\shortcite{Kim:14} and based on hyper parameters tested by Gehrmann \bgroup et al.\egroup~\shortcite{Gehrmann:17}.

The CNN model has one layer of convolution which used 4 different sized windows. Each window takes 2,3,4 or 5 words and applies 100 filters, encompassing the full embedding size.

We use this model to classify into Level 5 ICD-9 codes and first-level ICD-9 codes in the hierarchy.

\subsubsection{LSTM}

According to Yin \bgroup et al.\egroup~\shortcite{Yin:17} the state-of-the-art on many NLP tasks often switches between CNNs and RNNs (LSTM in this case), his paper lists different past studies where sometimes a CNN performs better and other times a LSTM.

Hence we implement a LSTM model to see if some of the discharge note features (e.g. temporal sequence) make it a better candidate. Since we have a relatevely small file (56K records), we start with a single layer LSTM to keep the number of parameters low. We didn't find published papers regarding classifying clinical notes using LSTM, however we did find a report on the web.\footnote{https://cs224d.stanford.edu/reports/priyanka.pdf}

\subsubsection{Attention}

As explored in Section~\ref{ss:preprocessing}, the average length of discharge clinical notes is 1639 words. The text to classify may be too long for a LSTM or CNN to remember all relevant information. 

Raffel et al.~\shortcite{Raffel:16} displayed better performance in many NLP tasks on long text using Attention. Here, we seek to emulate his results by implementing algorithms based on the formulas presented in ~\cite{Raffel:16} and Yang \bgroup et al.\egroup~\shortcite{Yang:16}.

{\bf LSTM with Attention}: The LSTM cell returns not only the last hidden state but all the intermediate ones that are then sent to the attention layer which creates a  new vector representing the clinical note for the output layer classification.

{\bf CNN with Attention}: The MaxPooling element in the CNN network is replaced by the Attention layer in order to create a vector representing all relevant information and not only taking in account max values. A model like this one is mentioned in Yin \bgroup et al.\egroup  ~\shortcite{Yin:16}

{\bf Hierarchical Attention}: This model was implemented based on Yang \bgroup et al.\egroup  ~\shortcite{Yang:16} which specifically targets document classifications. It has two levels of attention mechanisms, the first one creates vectors that represent each sentence, using attention mechanism across words; and the second level creates a vector that represent the document using attention mechanisms across sentences. Yang \bgroup et al.\egroup  ~\shortcite{Yang:16} uses Bidirectional GRUs while we use LSTMs for a fair comparision with the flat LSTM models.

\subsection{Threshold Calibration}

The resulting vector from the neural network may be interpreted as a probability of the individual ICD-9 codes (each cell has a value 0-1, but does not sum to 1). To complete the prediction, we must convert the vector to binary values.

There are several methods for selecting a Threshold ~\cite{Zhang:14}. We used a constant threshold maximizing the overall F1-score. In future work, we could explore methods building a (linear) model on top of the intermediate vector.

\subsection{Performance Metrics}

We use the F1 metric on the validation data to evaluate performance in all models and compare results with previous work on classifying MIMIC clinical notes and text classification in general.\footnote{For multi-label classifications, sklearn offers several options, we used the F1 'micro' option which calculates global counts for true positives, false negatives and false positives.}

\section{Results and Discussion}
\label{ssec:first}

\subsection{Comparing CNN with previous work}

In order to compare F1 performance results with the CNN model built by Gehrmann \bgroup et al.\egroup~\shortcite{Gehrmann:17}, we took into consideration the dataset size and number of classes.

Gehrmann's re-labeling approach is similar to our relabeling using the first-level ICD-9 codes in the ICD code hierarchy. Even though we have access to 52.6K records, we use a subset to relate to the 1.6K records used by Gehrmann \bgroup et al.\egroup~\shortcite{Gehrmann:17}. Since we have 17 classes, 7 more than the ones used by Gehrmann , we run our model with a dataset of 5K records.

Our CNN obtains similar result to Gehrmann \bgroup et al.\egroup, with a F1 score of 76.2\%, compare to their F1 score of 76\% (see Table~\ref{table:CNN_5k}). 

\begin{table}[h]

\begin{center}
\begin{tabular}{|p{1.6 cm}|p{1.2 cm}|p{1.4 cm}|l|l|}
\hline \bf Source & \bf Labels & \bf Methods & \bf Rec & \bf F1\\ \hline \hline
Gehrmann et al., 2017 & 10 own labels & LR 3-gram & 1.6K & 34.6 \\ \hline
Gehrmann et al., 2017 & 10 own labels & CNN & 1.6K & 76 \\ \hline
This Paper & 17 ICD-9 & CNN  & 5K & 76.2 \\
\hline
\end{tabular}
\end{center}
\caption{Classification of MIMIC clinical notes into labels representing high level phenotype categories (20 epochs for both CNN models)}
\label{table:CNN_5k}
\end{table}

\subsection{Testing CNN, LSTM and Attention}

To improve on this initial result, we ran experiments with the different models to identify the two more promising. These experiments run with a 5K notes, the 17 first level ICD-9 codes, using 5 epochs. The results are presented in Table ~\ref{table:attention}

We tested LSTMs with and without attention mechanisms, CNN with and without attention mechanisms and a Hierarchical LSTM model with Attention layers.

From the results in Table ~\ref{table:attention} we can see that CNN models do perform better than LSTM on classifying the MIMIC medical notes. 

\begin{table}[h]
\begin{center}
\begin{tabular}{|p{1.7 cm}|p{3 cm}|l|l|}
\hline \bf Source & \bf Methods & \bf Recs  & \bf F1 \\ \hline \hline
This Paper & LSTM&  5k &64.6 \\
This Paper & LSTM-Attention & 5k&  67 \\ \hline
This Paper & Hierarchical LSTM-Attention & 5k & 67.6 \\ \hline
This Paper & CNN& 5k&   69 \\
This Paper & CNN-Attention& 5k &  72.8 \\
\hline
\end{tabular}
\end{center}
\caption{Classification of MIMIC clinical notes into Level 1 ICD-9 Codes. Evaluation with 17 classes, 5k records, 5 epochs}
\label{table:attention}
\end{table}

We can also see that there is a significant improvement on F1 scores when applying attention mechanism to LSTM and CNN models.  The LSTM with Attention model outperforms the standard  LSTM by 2.4\% and the CNN with Attention model outperforms the standard CNN model by 3.8\%. 
 
On the other hand, the Hierarchical LSTM with Attention mechanisms had only a small increase (0.6\%) in performance results on regards to the Flat LSTM with Attention. This is smaller than we expected based on similar classification tasks by Yang \bgroup et al.\egroup~\shortcite{Yang:16}, where a difference of 3\% is reported, but on larger datasets. This model has twice the number of parameters than the flat models, which would impact performance for relatevely small files like the one we are using, this could be a reason for just a small   improvement in the f1 score.  We also tried GRUs instead of LSTMs to compare with Yang \bgroup et al.\egroup~\shortcite{Yang:16} results and the difference was still the same. 

Another possible reason our Hierarchical model is not performing much better is the tokenization of sentences. The model bases its predictions on the results on each sentence, and if the sentences are not identified correctly in the first place, then the rest of the model will not perform well. We did inspect suspicious long sentences which were not incorrect, they were lab reports. We would inspect closely the sentence tokenization process in further work. 

The two {\bf most promising models are CNN and CNN with Attention}, even the standard CNN model outperforms the Hierarchical model.

CNN models could be seen as hierarchical: the convolutional sliding windows create segments of the document (like sentences do) and they are collapsed into vectors representing a higher level of abstraction. In as sense CNN are finding the best segments in the document regardless of sentences separations. This may explain why the CNN models are getting a better performance than the Hierarchical models.

\subsection{CNN performance with full data set}

Here we show results from running the CNN models with the full data set.

First we classify clinical notes into the 20 most common Level 5 ICD-9 codes for comparison purposes: we can see that our model outperformed previous work (see Table \ref{table:top_20}).

\begin{table}[h]
\begin{center}
\begin{tabular}{|p{1.8 cm}|p {2.6 cm}|l|l|}
\hline \bf Source & \bf Methods & \bf N. Rec & \bf F1\\ \hline \hline
Perotte
et al., 2014& Hierarchal SVM (all codes) & 22K & 39.5 \\ \hline
Previous Project Reports\footnotemark& LSTM & 32K & 41.6 \\ \hline
This paper &Baseline & 46K & 35 \\ \hline
This paper &CNN & 46K & 72.4 \\
\hline
\end{tabular}
\end{center}
\caption{\label{font-table} Classification of MIMIC clinical notes into most common Level 5 ICD-9 Codes}
\label{table:top_20}
\end{table}
\footnotetext{https://cs224d.stanford.edu/reports/priyanka.pdf} 

To go further, we trained both CNN Models to classify clinical notes into Level 1 ICD-9 codes in the hierarchy (see Table~\ref{table:full_data}).

\begin{table}[h]
\begin{center}
\begin{tabular}{|p{1.8 cm}|l|l|l|}
\hline  \bf Source & \bf Methods & \bf Recs &\bf F1 \\ \hline \hline
This Paper & Baseline & 52.6K& 53 \\
This Paper &CNN & 52.6K& 79.7 \\
This Paper &CNN w/ Attention & 52.6K & 78.2 \\
\hline
\end{tabular}
\end{center}
\caption{\label{font-table} Classification of MIMIC clinical notes into 17 Level 1 ICD-9 Codes}
\label{table:full_data}
\end{table}

As anticipated earlier, the plain CNN model executed with the 52.6K records got a F1-score of 79.7\%, outperforming any other model in previous work, due to i) a larger dataset, ii) better separated labels, and iii) a more imbalanced label distribution (see Section~\ref{ss:labeling}).

However at this stage, the CNN ATT model still overfits: even though it had the highest score during the experimental runs with 5K records and 5 epochs, it didn't reach the best  f1-score when running it with the full data set. Further work would explore hyper-parameters tuning and evaluating the number of parameters to attempt undoing the over fitting situation.

%
%

We believe the CNN models can still be improved by inspecting in more detail cases where the model predicted a false positive or false negative, and working on hyper-parameters. This would be one of the first tasks to do in further work regarding these models.

\subsection{Pre-trained Embeddings}  

We used trainable embeddings, as described in Section~\ref{ss:embedding}. Our two attempts to initialize them with pre-trained values were unsuccessful.

Using the Wiki Glove pre-trained embeddings led to a minor decrease in performance (about 0.001\%) compared to an empty embedding matrix, which could be expected since Medical clinical notes have a vocabulary that differs from most Wiki pages. In fact, half of our vocabulary was not found on the Wiki Glove pre-trained embeddings.

We then created our own pre-trained embeddings using the Glove algorithm on all the MIMIC discharge notes. The result was a small performance improvement of 0.01\%.

As part of future work, we think that using pre-trained embeddings on millions of clinical notes would improve the performance of models processing clinical notes, this is an example of such type of work \footnote{https://web.stanford.edu/class/cs224n/reports/2744372.pdf}.

\section{Conclusions and outlook}

In this paper, we tested several alternative approaches for classifying ICD-9 codes. 

We showed that our a CNN models outperform significantly the F1 scores reported by previous work on Level 1 or Level 5 codes, while LSTMs and Hierarchical model displayed lower performance.

However for the problem of automatic labeling to be solved, models need to increase both in performance and in the precision of the codes that they allocate. Our results  highlight several areas to further that goal:

\begin{itemize}
\item optimization of CNN model with Attention, given promising results on small datasets
\item better adapt embeddings to clinical notes
\item broaden the number of ICD codes, gradually or by adapting the training metric
\end{itemize}

Finally we note that ICD codes are associated with a textual definition which could be directly compared with the clinical notes themselves.

\end{document}